\documentclass[runningheads]{llncs}
\pdfoutput=1
\usepackage{eccv}
\usepackage{eccvabbrv}
\usepackage{graphicx}
\usepackage{booktabs}
\usepackage{floatrow}
\newfloatcommand{capbtabbox}{table}[][\FBwidth]

\usepackage[accsupp]{axessibility}
\usepackage{hyperref}
\usepackage{orcidlink}

\begin{document}

\title{D$^4$-VTON: Dynamic Semantics Disentangling for Differential Diffusion based Virtual Try-On}
\titlerunning{D$^4$-VTON}
\author{Zhaotong Yang\inst{1} \and
Zicheng Jiang\inst{1} \and
Xinzhe Li\inst{1} \and
Huiyu Zhou\inst{2} \and
Junyu Dong\inst{1} \and
Huaidong Zhang\inst{3} \and
Yong Du\inst{1}\thanks{Corresponding author (csyongdu@ouc.edu.cn).}}
\authorrunning{Yang et al.}

\institute{Ocean University of China, Qingdao, China \and
University of Leicester, Leicester, United Kingdom \and
South China University of Technology, Guangzhou, China}

\maketitle

\begin{figure}
	\centering
    \includegraphics[width=0.98\linewidth]{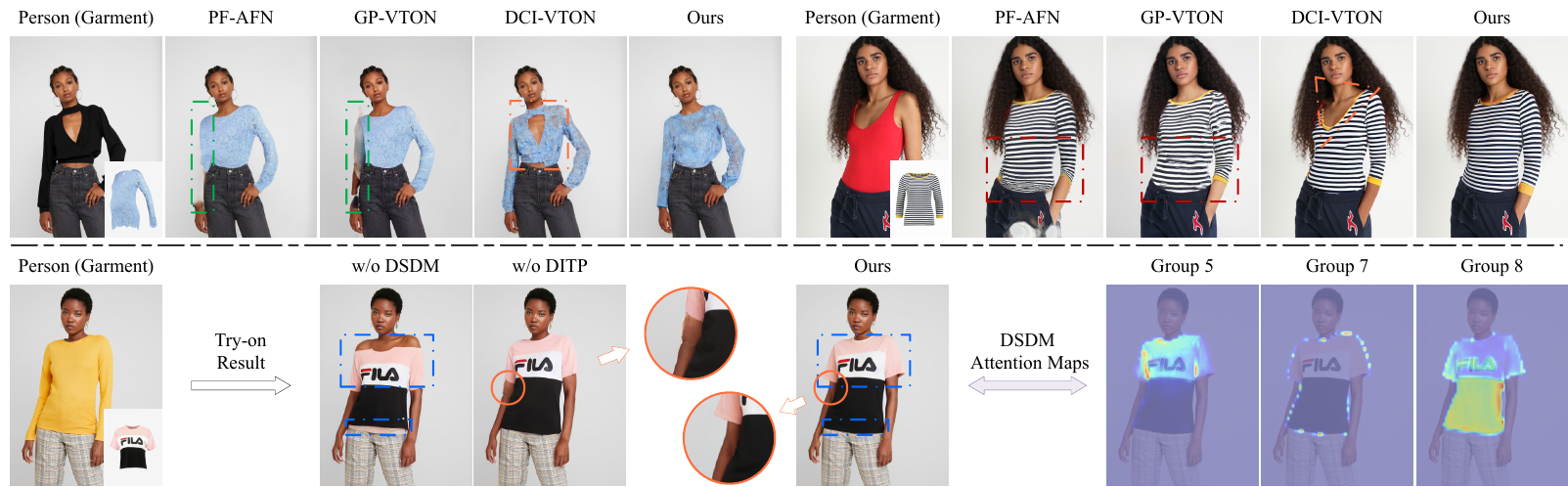}
    \caption{D$^4$-VTON excels with two innovations: i) Dynamic Semantics Disentangling Modules aggregate abstract semantic information for precise garment warping. ii) A diffusion-based framework integrating a Differential Information Tracking Path reduces learning ambiguities, enhancing accuracy in fitting garment types and body shapes.}
    \label{fig:teaser}
\end{figure}

\vspace{-10mm}\begin{abstract}
In this paper, we introduce D$^4$-VTON, an innovative solution for image-based virtual try-on. We address challenges from previous studies, such as semantic inconsistencies before and after garment warping, and reliance on static, annotation-driven clothing parsers. Additionally, we tackle the complexities in diffusion-based VTON models when handling simultaneous tasks like inpainting and denoising. Our approach utilizes two key technologies: Firstly, Dynamic Semantics Disentangling Modules (DSDMs) extract abstract semantic information from garments to create distinct local flows, improving precise garment warping in a self-discovered manner. Secondly, by integrating a Differential Information Tracking Path (DITP), we establish a novel diffusion-based VTON paradigm. This path captures differential information between incomplete try-on inputs and their complete versions, enabling the network to handle multiple degradations independently, thereby minimizing learning ambiguities and achieving realistic results with minimal overhead. Extensive experiments demonstrate that D$^4$-VTON significantly outperforms existing methods in both quantitative metrics and qualitative evaluations, demonstrating its capability in generating realistic images and ensuring semantic consistency. Code is available at \url{https://github.com/Jerome-Young/D4-VTON}.
\end{abstract}

\section{Introduction}\label{sec:intro}
With the rise of online shopping, image-driven virtual try-on (VTON) tasks have garnered significant interest due to their vast potential applications. The goal is to generate realistic try-on results given an image of clothing and a reference human image. This task hinges on two key aspects: first, accurately warping the clothing to align with the human body while preserving its appearance and texture details; and second, seamlessly integrating the warped clothing with the human figure to produce a lifelike result. Most existing methods~\cite{han2018viton,ge2021parser,fele2022c,he2022style,han2019clothflow,li2023virtual,xie2023gp} typically address these aspects in two stages: i) the warping stage, where the clothing is deformed to fit the body, and ii) the synthesis stage, where the incomplete human image is combined with the warped clothing to obtain the final try-on result.

Despite their advancements, current methods still have two primary limitations. Firstly, in the warping stage, they often use Thin Plate Splines (TPS)~\cite{bookstein1989principal} or appearance flows~\cite{zhou2016view} to warp garments. However, these techniques primarily focus on globally aligning clothing with the human body, overlooking local semantic variations in garment deformation. Specifically, they apply uniform deformation mappings across all clothing components, potentially sacrificing the fidelity of texture patterns (see the red box in Fig.~\ref{fig:teaser}). Recently, approaches~\cite{xie2023gp,li2023virtual,chen2023size} have sought to mitigate this issue by using clothing parsers to segment garments into semantic regions and learn distinct deformation mappings for each. However, training these parsers requires annotated semantics, which is both time-consuming and challenging to define appropriate semantic regions for precise deformation. Moreover, predictions made by the parsers are static during the garment deformation stage, hindering the correction of estimation errors.

Secondly, in the synthesis stage, methods typically employ generative models such as Generative Adversarial Networks (GANs)~\cite{goodfellow2014generative,Xu2021,Zhou2022,Song2022,Du2023} or diffusion models\cite{ho2020denoising,rombach2022high}. Unlike GANs~\cite{ge2021parser,xie2023gp,lee2022high,shim2023towards}, which can yield unrealistic outcomes (see the green box in Fig.~\ref{fig:teaser}), diffusion models leverage conditional guidance, provide more stable training, and integrate broader domain knowledge for generating high-quality results. However, VTON methods using diffusion models~\cite{morelli2023ladi,gou2023taming,li2023virtual} face a complex task in this phase, necessitating simultaneous optimization of the model's denoising and inpainting processes for effective try-on results. Existing approaches often lack specific objectives tailored for these tasks. For instance, DCI-VTON~\cite{gou2023taming} utilizes distinct inputs to train denoising and inpainting processes separately within a shared model but relies on inherently similar guidance from ground truth images for both tasks. This approach may introduce learning ambiguities that impact the accurate reconstruction of synthesis results, including garment (see the orange box in Fig.~\ref{fig:teaser}) and human body reconstructions (shown at the bottom of Fig.~\ref{fig:teaser}). Hence, there is a pressing need to pioneer a new virtual try-on paradigm anchored in diffusion models.

In this paper, we present D$^4$-VTON, a novel $\textbf{V}$irtual $\textbf{T}$ry-$\textbf{ON}$ solution that combines a \textbf{D}ynamic semantics \textbf{D}isentangling technique and a \textbf{D}ifferential \textbf{D}iffu-sion based framework to address the above issues. Inspired by previous studies on segment-level grouping~\cite{Xu2021a,tang2023contrastive}, which assume similar semantic information is shared among a subset of channels in feature maps, our disentangling approach leverages the self-similarity of garment features to dynamically form distinct groups of abstract semantics. This enables the independent learning of local flows for each group, guiding garment deformation in a semantically disentangled manner beyond reliance on clothing parsers. Consequently, it enhances the accuracy of pattern deformation while achieving global alignment of the garment with the human body. 

Additionally, our diffusion framework introduces a differential information tracking path that monitors the difference between complete and incomplete noisy inputs during the synthesis stage. Unlike previous methods that jointly restore multiple degradations (\ie, denoising and inpainting), our approach separates these processes. By utilizing differential information, we first fill in the incomplete noisy input and then denoise the completed result, thus alleviating the learning ambiguities caused by simultaneous optimization. Comprehensive experiments demonstrate that D$^4$-VTON surpasses state-of-the-art methods both quantitatively and qualitatively on several benchmarks, showcasing its superiority in generating realistic try-on results with precise semantic consistency.

In summary, our key contributions are as follows:
\begin{itemize}
	\item We present D$^4$-VTON, a virtual try-on model that effectively preserves garment texture patterns after warping and produces high-fidelity try-on results through an advanced synthesis process. 
	\item We introduce dynamic semantics disentangling modules (DSDMs) that leverages the self-similarity of garment features to independently learn local flows, enabling semantically disentangled garment deformation.
	\item We propose a novel diffusion paradigm for VTON that minimizing learning ambiguities caused by multi-degradation restoration and enhances the synthesis performance by differential information tracking. 
	\item We surpass state-of-the-art methods by a significant margin on multiple benchmarks quantitatively, while also demonstrating the realism and accuracy of our try-on results in qualitative evaluations.
\end{itemize}
\section{Related Work}
\textbf{Image-based Virtual Try-on.} 
To improve the accuracy of try-on results, most existing VTON methods~\cite{han2018viton,ge2021parser,he2022style,li2023virtual,xie2023gp,bai2022single,lee2022high,han2019clothflow,fele2022c,xie2021towards} focus on designing robust warping networks. For example, VITON~\cite{han2018viton} and CP-VTON~\cite{Wang2018} use Thin Plate Splines (TPS) to fit garment deformations, but TPS relies on sparse point correspondences, making it inadequate for complex poses and occlusions. ClothFlow~\cite{han2019clothflow} addresses this by introducing appearance flow, which predicts dense pixel correspondences to better capture complex deformations. ACGPN~\cite{Yang2020} preserves unchanged regions by replacing the clothing-agnostic image with a semantic segmentation map. PF-AFN~\cite{ge2021parser} reduces reliance on parsers through knowledge distillation, and StyleFlow~\cite{he2022style} employs the StyleGAN~\cite{karras2019style} architecture to capture global contextual information. However, these methods primarily focus on geometrically aligning garments with the body, neglecting local garment pattern deformations, leading to texture distortion. In contrast, our method introduces DSDMs to independently learn local flows for different semantics, achieving more accurate garment deformation, particularly in texture details.\\
\textbf{Segment-level Grouping Strategies.} Segment-level grouping strategies are crucial for manipulating different parts of an image, widely used in object detection, semantic segmentation, and image synthesis. For instance, Grouplane~\cite{li2023grouplane} uses group detection for one-to-one matching with ground truth in lane line prediction, while CGFormer~\cite{tang2023contrastive} captures instance-level information via grouping tokens for referring image segmentation. iPOSE~\cite{wei2023inferring} learns object part maps through a few-shot regime for semantic image synthesis. However, these methods rely on supervised information, which is time-consuming to collect for virtual try-on tasks. 

\vspace{-0.5mm}In virtual try-on, GP-VTON~\cite{xie2023gp} deforms clothing by independently warping the left sleeve, right sleeve, and torso to address adhesive issues. KGI~\cite{li2023virtual} and COTTON~\cite{chen2023size} further divide upper garments into five parts—left lower, left upper, middle, right lower, and right upper—for more precise deformation. These methods depend on parsers for grouping, introducing unavoidable errors and restricting group granularity due to limited semantic annotations. In contrast, our DSDM captures abstract semantics in a self-discovered manner, enhancing clothing deformation realism without relying on parsers.\\
\textbf{Diffusion Models.} The popularity of diffusion models stems from the introduction of the denoising diffusion probabilistic model (DDPM)~\cite{sohl2015deep,ho2020denoising}, which hypothesize two Markov chains: one progressively adding noise to an image until it approximates a Gaussian distribution, and the other reversing this process. However, the slow sampling speed of DDPMs limits their application. DDIM~\cite{song2020denoising} addresses this by transforming the sampling process into a non-Markovian one to accelerate it. To reduce the computational burden of training with high-resolution images, LDM~\cite{rombach2022high} encodes images into latent space and uses conditional guidance to generate controllable, diverse results.

Diffusion-based virtual try-on methods have surpassed GAN-based methods, becoming mainstream due to their high-quality, realistic outputs. KGI~\cite{li2023virtual} uses a content-keeping mask to preserve realistic textures of deformed clothing, but this can cause ``copy and paste'' artifacts at the clothing-body boundaries. LADI-VTON~\cite{morelli2023ladi} employs textual inversion for better accuracy, yet textual information often fails to capture specific clothing patterns. DCI-VTON~\cite{gou2023taming} introduces a dual-branch diffusion model for denoising and inpainting incomplete noisy inputs simultaneously. However, the differing inputs and intrinsically similar objectives of the branches can lead to optimization conflicts. In contrast, our method uses differential information tracking to separately recover multiple degradations with distinct objectives for different inputs, reducing learning ambiguities.
\section{Methods}\label{sec:Methods}\setcounter{footnote}{0}
\begin{figure}\vspace{-3mm}
	\centering
	\includegraphics[width=\textwidth]{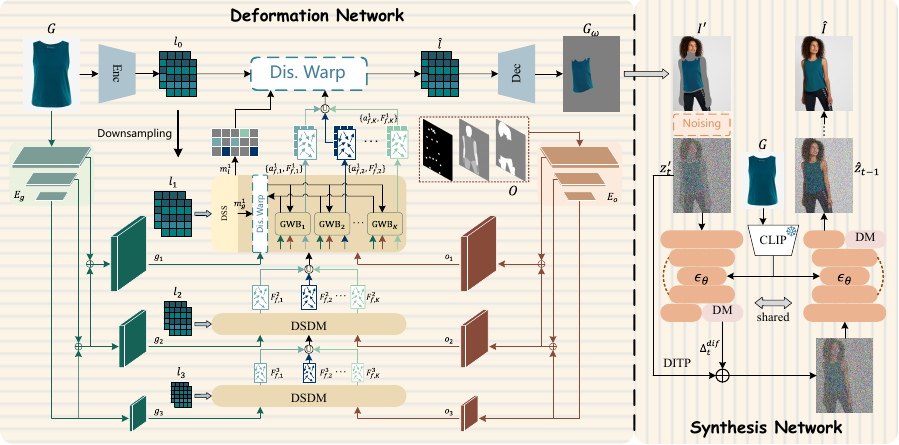}
	\caption{Overall pipeline of D$^4$-VTON. The deformation network takes the garment image $G$ and conditional triplet $O$ to generate local flows via DSDMs. Utilizing the final flows, we warp the garment features $l_0$ and decode them into the warped garment $G_\omega$, which is then combined with the clothing-agnostic image to create $I^\prime$ as the input for the synthesis network. By tracking differential information via DITP, we separately perform inpainting and denoising on the latents to produce the try-on result $\hat{I}$. }
	\label{fig:framework}	
\end{figure}
We adopt a two-stage framework for the virtual try-on task with two main goals: 1) to disentangle garment semantics in a self-discovered manner and learn local flows for each, and 2) to design a new diffusion paradigm to avoid learning ambiguities in the synthesis stage. Fig.~\ref{fig:framework} illustrates the overall pipeline of the proposed D$^4$-VTON.

In our deformation network, we first take the garment image $G$ and the condition triplet $O$ (human pose, densepose pose, and preserve region mask) as input. We use two feature pyramid networks (FPN)~\cite{lin2017feature}, $E_g(\cdot)$ and $E_o(\cdot)$, to extract multi-scale features $\{g_1,g_2,\cdots,g_N\}=E_g(G)$ and $\{o_1,o_2,\cdots,o_N\}=E_o(O)$\footnote{For simplicity, we illustrate the case with $N$=3 in Fig.~\ref{fig:framework}.}. These features, along with garment features $\{l_1,l_2,\cdots,l_N\}$ from a lightweight encoder $\phi(\cdot)$, where $l_i=\phi(D(G,i))$ and $D(G,i)$ denotes the $i$-th downsampled result of $G$, are processed by our Dynamic Semantics Disentangling Module (DSDM) to generate local flows and attention maps from coarse to fine. The final predicted flows and attention maps warp the garment features $l_0$ and decode them into the warped garment $G_w$. It is then combined with the clothing-agnostic image to obtain $I^\prime$ as input for the synthesis network. By leveraging differential information tracking, we avoid learning ambiguities by separately performing denoising and inpainting tasks in the reverse process of the diffusion model, rather than restoring multiple degradations simultaneously, to generate the final try-on result $\hat{I}$.\\
\vspace{-6mm}\subsection{Dynamic Semantics Disentangling Module}\label{sec:DGWM}
\begin{figure}\vspace{-4mm}
	\centering
	\includegraphics[width=\textwidth]{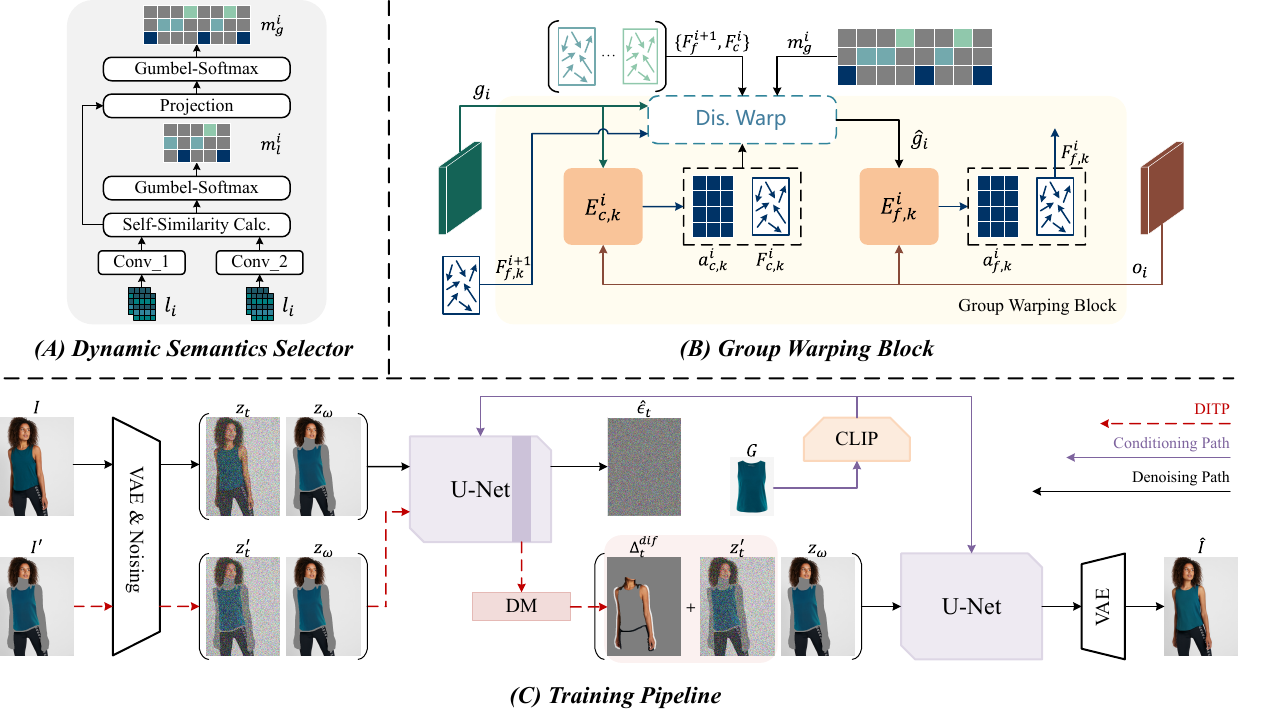}
	\caption{Illustration of the Dynamic Semantics Selector, the Group Warping Block, and the training pipeline of the differential diffusion based synthesis network.}
	\label{fig:detail}
\end{figure}
Our DSDM consists of three main components: 1) Dynamic Semantics Selector, which generates masks to dynamically select and aggregate feature channels with similar semantics; 2) Group Warping Block, responsible for generating flows and attention maps for different channel groups; and 3) Disentangled Warping Operation, which uses the selection mask and predicted flows and attention maps to deform garment features.\\
\textbf{Dynamic Semantics Selector.} To dynamically select feature channels that capture similar semantics, we use the self-similarity of garment features to unsupervisedly generate selection masks indicating each channel's semantic group. Specifically, for the feature $l_i\in\mathbb{R}^{Q\times H\times W}$ at the $i$-th scale, with a predefined number of groups $K$, we pass it through two convolutional layers, $\textrm{Conv}\_1(\cdot)$ and $\textrm{Conv}\_2(\cdot)$, using $K$ and $Q$ kernels of size $3\times3$ respectively, to obtain two sets of features from the garment. We then flatten these two features into matrices: $S^1_i\in\mathbb{R}^{K\times HW}$, which represents $K$ abstract semantics, and $S^2_i\in\mathbb{R}^{Q\times HW}$, capturing the garment features. Subsequently, we can obtain the self-similarity matrix $S_{self}^i\in\mathbb{R}^{K\times Q}$, which can be calculated as
\vspace{-1mm}\begin{equation}
	S_{self}^i=S^1_i\cdot S^2_i.
\end{equation}
In matrix $S_{self}^i$, each element indicates how each channel responds to a specific semantic. We determine the semantic group for each channel by filtering based on these responses. This operation employs Gumbel-Softmax~\cite{jang2016categorical,maddison2016concrete}, a differentiable method ensuring exclusive assignment of each channel to a distinct group. This process can be formulated as follows:
\begin{equation}
	\begin{aligned}
		&m_{soft}^i=\mathrm{softmax}((S_{self}^i+A_{gumbel})/\mathop{\tau}\limits_{\ }), \\
		&m_{hard}^i=\mathrm{onehot}(\mathop{\arg\max}_{K}(m_{soft}^i)), \\
		&m_l^i=(m_{hard}^i)^\top-\mathrm{sg}(m_{soft}^i)+m_{soft}^i, \\
	\end{aligned}
\end{equation}
where $A_{gumbel}\in\mathbb{R}^{K\times Q}$ is sampled from a $\mathrm{Gumbel}(0,1)$ distribution, and $\tau$ is a learnable temperature parameter. The operation $\mathop{\arg\max}_{K}(\cdot)$ selects the maximum element value along each column of the $K\times Q$ matrix. The $\mathrm{onehot}(\cdot)$ function converts the soft distribution into $Q$ one-hot vectors, and $\mathrm{sg}(\cdot)$ denotes the stop gradient operation. Note that besides using the selection mask $m_l^i$ during the decoding of warped features, we also need a selection mask within each group warping block to facilitate learning local flows in a coarse-to-fine manner. To ensure dimensional compatibility in computations, our selector adjusts $S_{self}^i$ to match the required dimensions before applying the Gumbel-Softmax operation to derive the mask $m_g^i$. Fig.~\ref{fig:detail}(A) illustrates the specifics of the selector.\\
\textbf{Group Warping Block.} Our Group Warping Block (GWB) cascades a coarse estimator $E_c(\cdot)$ and a fine estimator $E_f(\cdot)$ for coarse/fine flows estimation, as illustrated in Fig.~\ref{fig:detail}(B). Specifically, for the features $g_i$ and $o_i$ at the $i$-th scale, we first compute the coarse flow $F_c^i$ and the attention map $a_c^i$ as follows:
\begin{equation}
	\{F_c^i,a_c^i\}=E_c^i(g_i,o_i).
	\label{equ:enccoarse}
\end{equation}
Next, we use the coarse and fine flows $F_c^{i}, F_f^{i+1}$, the attention map $a_c^i$, and the selection mask $m_g^i$ to warp the garment features $g_i$ in a disentangled manner, formulated as follows:
\begin{equation}
	\hat{g}_i=\mathcal{W}_{dis}(F_c^{i}, F_f^{i+1},a_c^i,m_g^i,g_i),
\end{equation}
where $\mathcal{W}_{dis}(\cdot)$ denotes our disentangled warping operation, which will be explained later. Similar to Eq.~(\ref{equ:enccoarse}), we then use the warped features $\hat{g}_i$ and $o_i$ to obtain the local fine flow $F_f^i$ and the attention map $a_f^i$:
\begin{equation}
	\{F_f^i,a_f^i\}=E_f^i(\hat{g}_i,o_i),
	\label{equ:encfine}
\end{equation}
Note that each GWB is used for a local flow estimation corresponding to a specific group of semantic, and thus a DSDM contains $K$ such blocks.\\
\textbf{Disentangled Warping Operation.} Now we elaborate on how $\mathcal{W}_{dis}(\cdot)$ deforms the features. This operation is used in two places: first, within the GWB to generate intermediate warped features $\hat{g}_i$ for obtaining local fine flows; second, for generating warped garment features used for decoding.

For the former case, we first use $F_c^i$ to warp the fine flow $F_f^{i+1}$ generated by the corresponding GWB from the previous level, that is
\begin{equation}
	F_c^{i\prime}=\mathcal{G}(F_c^i, F_f^{i+1}),
	\label{equ:warp1}	
\end{equation} 
where $\mathcal{G}(\cdot)$ denotes the grid-sampling operator. Consequently, $F_c^{i\prime}$ can integrate flow information from multi-scale features. Note that this operation is not performed at the $N$-th scale. 

Next, we aggregate the features from different groups using their corresponding flow, attention map, and their respective group indicators (\ie, the corresponding row of the selection mask) to form intermediate warped features. For the $i$-th scale, this process can be formulated as follows:
\begin{equation}
	\hat{g}_i=\sum_{j=1}^{K}\frac{\textrm{exp}(a_c^j)}{\sum_{k=1}^{K}\textrm{exp}(a_c^k)}\mathcal{G}(F_{c,j}^{i\prime},g_i\odot m_g^i(j,:)),
	\label{equ:warp2}
\end{equation}
where $\odot$ denotes the replication and dot product operation. This allows the warping operation to apply different local flows for different semantics.

On the other hand, the warped garment features $\hat{l}_i$, used for decoding, are obtained through a process similar to that in Eq.~(\ref{equ:warp1}) and (\ref{equ:warp2}). The only difference is that here we use $F_f^i$ to warp $F_c^i$ first, and it can be formulated as follows:
\begin{equation}
	\hat{l}_i=\mathcal{W}_{dis}(F_f^i,F_c^i,a_f^i,m_l^i,l_i).
\end{equation}
\textbf{Learning Objectives.} We optimize the deformation network by comparing the similarity between the generated warped garment at all scales and their corresponding ground truth. Since flow deformation occurs in the feature space, we obtain smooth results without the need for any constraints or regularization on the flow~\cite{bai2022single}. The total learning objective including three terms: $\mathcal{L}_{mae}$, $\mathcal{L}_{prec}$, and $\mathcal{L}_{style}$, which is defined as follows:
\begin{equation}
	\mathcal{L}_{def}=\mathcal{L}_{mae}+\mathcal{L}_{prec}+\lambda_{style}\mathcal{L}_{style},
\end{equation}
where $\lambda_{style}$ is a balance factor. Specifically, $\mathcal{L}_{mae}$ is defined as follows:
\begin{equation}
	\mathcal{L}_{mae}=\Vert{G_\omega - I_\omega}\Vert_1,
\end{equation}
where $\Vert\cdot\Vert_1$ denotes the $L_1$ distance, and $I_\omega$ indicates the ground truth warped garment image. $\mathcal{L}_{prec}$ computes the $L_1$ distance of the feature maps using VGG-19~\cite{simonyan2014very}:
\begin{equation}
	\mathcal{L}_{prec}=\sum_{i}\Vert\Phi_i(G_\omega)-\Phi_i(I_\omega)\Vert_1,
\end{equation}
where $\Phi_i(\cdot)$ represents the feature map at the $i$-th layer of VGG-19 $\Phi(\cdot)$. Additionally, $\mathcal{L}_{style}$ is calculated as follows:
\begin{equation}
	\mathcal{L}_{style}=\sum_i\Vert \textrm{Gram}_{\Phi_i}(G_\omega)-\textrm{Gram}_{\Phi_i}(I_\omega)\Vert_1,
\end{equation}
where $\textrm{Gram}_{\Phi_i}(\cdot)$ denotes the Gram matrix of the feature maps from the $i$-th layer of $\Phi(\cdot)$.
\subsection{Differential Diffusion based VTON}\label{sec:DNRP}
After obtaining the warped garment, D$^4$-VTON uses a diffusion model to synthesize the try-on result. Images offer precise details and spatial information lacking in text conditions, prompting our synthesis network to adopt the architecture of PBE~\cite{yang2023paint} with its pretrained parameters. PBE, trained on millions of images, excels in image inpainting with similar semantics and styles conditioned on exemplar images. However, for virtual try-on tasks, precisely reconstructing garment patterns solely from exemplar images proves inadequate~\cite{gou2023taming}. To address this, we propose a novel diffusion pipeline featuring two paths: one for denoising and another for inpainting.

The denoising path takes the ground truth human image $I$ as input, while the inpainting path utilizes $I^\prime$, created by integrating the warped garment into the clothing-agnostic image. Notably, $I^\prime$ is also incorporated as a condition concatenated with both inputs at each training step to guide generation. However, the standard objective, which measures the similarity between predictions (noises or clean and completed images) and ground truths, conflicts with simultaneously managing these two distinct tasks—denoising and inpainting—across different paths, potentially causing learning ambiguities. Therefore, we introduce a Differential Information Tracking Path (DITP) to replace a simplistic inpainting path. Along this path, the objective is not to reconstruct noises or clean and completed images, but rather to understand the distinctions between inputs from both paths. This approach allows us to initially complete $I^\prime$ in the latent space using these distinctions, followed by separate denoising of the resulting latents, effectively addressing multiple degradations and mitigating learning ambiguities. Fig.~\ref{fig:detail}(C) outlines the training pipeline of our proposed diffusion framework.\\
\textbf{Denoising Path.} Similar to the training method of vanilla LDM, the denoising path begins by initiating the diffusion process on a given human image $I$, resulting in the latent code $z_t$ at the $t$-th timestep using a pre-trained VAE encoder~\cite{kingma2013auto}. Subsequently, this latent code $z_t$ is fed into the U-Net architecture $\epsilon_\theta(\cdot)$, alongside the latent code $z_\omega$ corresponding to $I^\prime$. Guided by the global condition $q=\textrm{CLIP}(G)$, $\mathcal{L}_{dng}$ directs the learning process by quantifying the $L_2$ distance between the prediction and the ground truth noise $\epsilon$:
\begin{equation}
	\mathcal{L}_{dng}=\mathbb{E}_{z_0,z_\omega,M,q,t}[\Vert\epsilon-\epsilon_\theta(z_t,z_\omega,M,q,t)\Vert_2],
\end{equation}
where $M$ denotes the downsampled inpainting mask adjusted to match the latent dimensions.\\
\textbf{Differential Information Tracking Path.} We effectively capture the differential information between the latent code $z_t$ and its incomplete counterpart $z_t^\prime$ using DITP, with minimal increase in parameter count. Specifically, we introduce a lightweight differential mapper (DM) deployed just before the output layer of the last U-Net block to predict the differential information $\mathrm{\Delta}_t^{dif}$. This is guided by a simple loss term $\mathcal{L}_{dif}=\Vert z_t-z_t^\prime\Vert_2$. In addition to addressing the issue of restoring multiple degradations jointly, DITP enables us to generate an extra set of inputs $\{\mathrm{\Delta}_t^{dif}+z_t^\prime,z_\omega\}$ for the denoising path, ensuring homogeneity of the training data under the same reconstruction loss, while simultaneously enriching the denoising path's training dataset.

Finally, the overall objective $\mathcal{L}_{syn}$ for our synthesis network is defined as follows:
\begin{equation}
	\mathcal{L}_{syn}=\mathcal{L}_{rec}+\gamma_{dif}\mathcal{L}_{dif}+\gamma_{MAE}\mathcal{L}_{MAE}+\gamma_{PREC}\mathcal{L}_{PREC},
\end{equation}
where $\mathcal{L}_{rec}=\Vert\epsilon-\epsilon_\theta(z_t,\mathrm{\Delta}_t^{dif}+z_t^\prime,z_\omega,M,q,t)\Vert_2$ now redefines $\mathcal{L}_{dng}$. $\mathcal{L}_{MAE}$ and $\mathcal{L}_{PREC}$ are based the try-on result $\hat{I}$ similar to previous work~\cite{gou2023taming}, and $\gamma_{dif}$, $\gamma_{MAE}$, and $\gamma_{PREC}$ are hyperparameters.

\section{Experiments}\label{sec:Experiments}
\textbf{Datasets.} We evaluate the proposed method on two most popular datasets, VITON-HD~\cite{choi2021viton} and DressCode~\cite{morelli2022dress}. The VITON-HD dataset is collected for high-resolution virtual try-on tasks, consisting of 11,647 image pairs in the training set and 2,032 pairs in the testing set. The DressCode dataset includes three categories—dress, upper, and lower—with a combined total of 48,392 training image pairs and 5,400 testing pairs. Both datasets include in-shop garment images and corresponding ground truth human images.\\
\textbf{Baselines and Evaluation metrics.} We compared our method against various state-of-the-art approaches, including CNN-based (SDAFN~\cite{bai2022single}), GAN-based (PF-AFN~\cite{ge2021parser}, FS-VTON~\cite{he2022style}, HR-VTON~\cite{lee2022high}, GP-VTON~\cite{xie2023gp}, SD-VTON~\cite{shim2023towards}), and diffusion-based methods (LADI-VTON~\cite{morelli2023ladi}, DCI-VTON~\cite{gou2023taming}). GP-VTON and SD-VTON results were obtained from their official checkpoints, while the others were retrained using their official codes to achieve results at a resolution of 512$\times$384.

We conducted evaluations using two testing settings: paired, where a garment image reconstructs the image of the person originally wearing it, and unpaired, which involves changing the garment worn by a person's image. For both evaluation settings, we utilized two widely-used metrics: Frechet Inception Distance (FID)~\cite{parmar2022aliased} and Kernel Inception Distance (KID)~\cite{binkowski2018demystifying}. Additionally, in the paired setting where ground truth is available for comparison, we included Peak Signal-to-Noise Ratio (PSNR), Structural Similarity (SSIM)\cite{wang2004image}, and Learned Perceptual Image Patch Similarity (LPIPS)\cite{zhang2018unreasonable} to assess the accuracy of the try-on results. Furthermore, we incorporated user studies to capture human perception, providing a comprehensive comparison across all evaluated methods.\\
\begin{table}
	\centering
	\def\arraystretch{1.2}
	\small
	\scriptsize
	\tabcolsep 1.5pt
	\resizebox{0.78\textwidth}{!}{
		\begin{tabular}{ll ll lllllll}
			\toprule
			\multicolumn{2}{l}{Method} && FID$_u$$\downarrow$ & KID$_u$$\downarrow$ && FID$_p$$\downarrow$ & KID$_p$$\downarrow$ & PSNR$_p$$\uparrow$ & SSIM$_p$$\uparrow$ & LPIPS$_p$$\downarrow$ \\
			\midrule
			\multicolumn{2}{l}{PF-AFN~\cite{ge2021parser}} && 9.654 & 1.04 && 6.554 & 0.81 & 23.54 & \underline{0.888} & 0.087 \\
			\multicolumn{2}{l}{FS-VTON~\cite{he2022style}} && 9.908 & 1.10 && 6.170 & 0.69 & 23.79 & 0.886 & \underline{0.074}\\
			\multicolumn{2}{l}{HR-VTON~\cite{lee2022high}} && 13.265 & 4.38 && 11.383 & 3.52 & 21.61 & 0.865 & 0.122 \\
			\multicolumn{2}{l}{SDAFN~\cite{bai2022single}} && 9.782 & 1.11 && 6.605 & 0.83 & 23.24 & 0.880 & 0.082 \\
			\multicolumn{2}{l}{GP-VTON~\cite{xie2023gp}} && 9.365 & 0.79 && 6.031 & 0.60 & 23.41 & 0.885 & 0.080 \\
			\multicolumn{2}{l}{LADI-VTON~\cite{morelli2023ladi}} && 9.346 & 1.66 && 6.602 & 1.09 & 22.49 & 0.866 & 0.094 \\
			\multicolumn{2}{l}{DCI-VTON~\cite{gou2023taming}} && \underline{8.754} & \underline{0.68} && \underline{5.521} & \underline{0.41} & \underline{24.01} & 0.882 & 0.080\\
			\multicolumn{2}{l}{SD-VTON~\cite{shim2023towards}} && 9.846 & 1.39 && 6.986 & 1.00 & 22.73 & 0.874 & 0.101 \\
			\cmidrule{1-2} \cmidrule{4-5} \cmidrule{7-11} 
			\multicolumn{2}{l}{Ours} && \textbf{8.530} & \textbf{0.25} && \textbf{4.845} & \textbf{0.04} & \textbf{24.71} & \textbf{0.892} & \textbf{0.065} \\
			\bottomrule
		\end{tabular}
	}	
	\caption{Quantitative comparisons on VITON-HD dataset~\cite{choi2021viton}. Subscripts $u$ and $p$ denote the unpaired and paired settings, respectively. The best results are highlighted in \textbf{bold} and the second best are \underline{underlined}.}
	\label{tab:vitonhd}
\end{table}
\begin{table}
	\centering
	\def\arraystretch{1.2}
	\scriptsize
	\tabcolsep 1.5pt
	\resizebox{\textwidth}{!}{
		\begin{tabular}{l l l l l l l l l l l l l l l l l l l l}
			\toprule
			\multicolumn{2}{c}{Dataset}& & \multicolumn{5}{c}{DressCode-Upper} & & \multicolumn{5}{c}{DressCode-Lower} & & \multicolumn{5}{c}{DressCode-Dresses} \\
			\cmidrule{1-2} \cmidrule{4-8} \cmidrule{10-14} \cmidrule{16-20}
			\multicolumn{2}{l}{Method}                              
			& & FID$_{u/p}$ $\downarrow$ & KID$_{u/p}$ $\downarrow$ & PSNR $\uparrow$ & SSIM $\uparrow$ & LPIPS $\downarrow$
			& & FID$_{u/p}$ $\downarrow$ & KID$_{u/p}$ $\downarrow$ & PSNR $\uparrow$ & SSIM $\uparrow$ & LPIPS $\downarrow$
			& & FID$_{u/p}$ $\downarrow$ & KID$_{u/p}$ $\downarrow$ & PSNR $\uparrow$ & SSIM $\uparrow$ & LPIPS $\downarrow$ \\
			\cmidrule{1-2} \cmidrule{4-8} \cmidrule{10-14} \cmidrule{16-20}
			
			\multicolumn{2}{l}{PF-AFN~\cite{ge2021parser}} 
			& & 17.581/7.384 & 6.66/1.98 & 26.87 & 0.940 & 0.037
			& & 19.857/9.541 & 8.32/2.54 & 26.55 & 0.930 & 0.056
			& & 19.824/11.269 & 8.56/3.26 & 23.80 & 0.885 & 0.074 \\
			\multicolumn{2}{l}{FS-VTON~\cite{he2022style}} 
			& & 16.342/11.293 & 5.93/3.65 & 26.32 & 0.941 & \underline{0.035}
			& & 22.432/11.652 & 9.81/3.82 & 26.38 & 0.934 & 0.053
			& & 20.950/13.044 & 8.96/4.44 & 22.18 & \underline{0.888} & \underline{0.070} \\
			\multicolumn{2}{l}{HR-VITON~\cite{lee2022high}} 
			& & 16.824/15.365 & 5.70/5.27 & 21.92 & 0.916 & 0.071
			& & 16.392/11.409 & 4.31/3.20 & 25.96 & 0.937 & 0.045
			& & 18.812/16.821 & 5.41/4.89 & 21.06 & 0.865 & 0.113 \\
			\multicolumn{2}{l}{SDAFN~\cite{bai2022single}} 
			& & 11.734/8.053 & 1.09/\underline{0.67} & 27.01 & 0.936 & 0.039
			& & 16.089/\underline{7.144} & 3.07/\underline{0.55} & \underline{27.24} & 0.933 & 0.049
			& & 12.548/\underline{7.915} & \underline{1.27}/\underline{0.45} & 24.05 & 0.879 & 0.082 \\
			\multicolumn{2}{l}{GP-VTON~\cite{xie2023gp}} 
			& & 12.208/\underline{7.380} & 1.19/0.74 & 26.48 & \underline{0.945} & 0.036
			& & 16.700/7.733 & 2.89/0.71 & 25.23 & 0.938 & \underline{0.042}
			& & 12.643/\textbf{7.442} & 1.83/\textbf{0.32} & 22.56 & 0.881 & 0.073 \\
			\multicolumn{2}{l}{DCI-VTON~\cite{gou2023taming}} 
			& & \underline{11.640}/7.466 & \underline{0.86}/1.07 & \underline{27.14} & 0.942 & 0.041
			& & \underline{15.450}/7.969 & \underline{1.60}/0.96 & 26.61 & \underline{0.939} & 0.045
			& & \underline{12.348}/8.484 & 1.36/1.08 & \underline{24.07} & 0.887 & \underline{0.070} \\
			\cmidrule{1-2} \cmidrule{4-8} \cmidrule{10-14} \cmidrule{16-20}
			\multicolumn{2}{l}{\textbf{Ours}} 
			& & \textbf{10.995}/\textbf{6.548} & \textbf{0.30}/\textbf{0.23} & \textbf{27.25} & \textbf{0.946} & \textbf{0.032}
			& & \textbf{14.855}/\textbf{6.677} & \textbf{1.50}/\textbf{0.04} & \textbf{27.62} & \textbf{0.946} & \textbf{0.033}
			& & \textbf{12.289}/8.276 & \textbf{1.14}/1.07 & \textbf{24.48} & \textbf{0.890} & \textbf{0.061} \\  
			\bottomrule
		\end{tabular}
	}	
	\caption{Quantitative comparisons on DressCode dataset~\cite{morelli2022dress}. Subscripts $u$ and $p$ denote the unpaired and paired settings, respectively. The best results are highlighted in \textbf{bold} and the second best are \underline{underlined}.}\vspace{-2mm}
	\label{tab:dresscode}
\end{table}
\textbf{Quantitative Results.} In Tab.~\ref{tab:vitonhd}, we present the quantitative results on the VITON-HD dataset compared to all comparisons. For the paired setting, D$^4$-VTON surpasses the second-best method by 0.70dB and 0.004 on PSNR and SSIM, respectively, indicating more accurate results for the warped garments and the try-on outcomes. More importantly, our method demonstrates strong generality in the unpaired setting, with improvements of 0.224 in FID$_u$ and 0.43 in KID$_u$ over the second-best method. Additionally, D$^4$-VTON shows significant improvements in FID$_p$, KID$_p$ and LPIPS metrics, further highlighting its advanced performance in terms of perceptual quality.

\begin{figure}
	\centering
	\includegraphics[width=\textwidth]{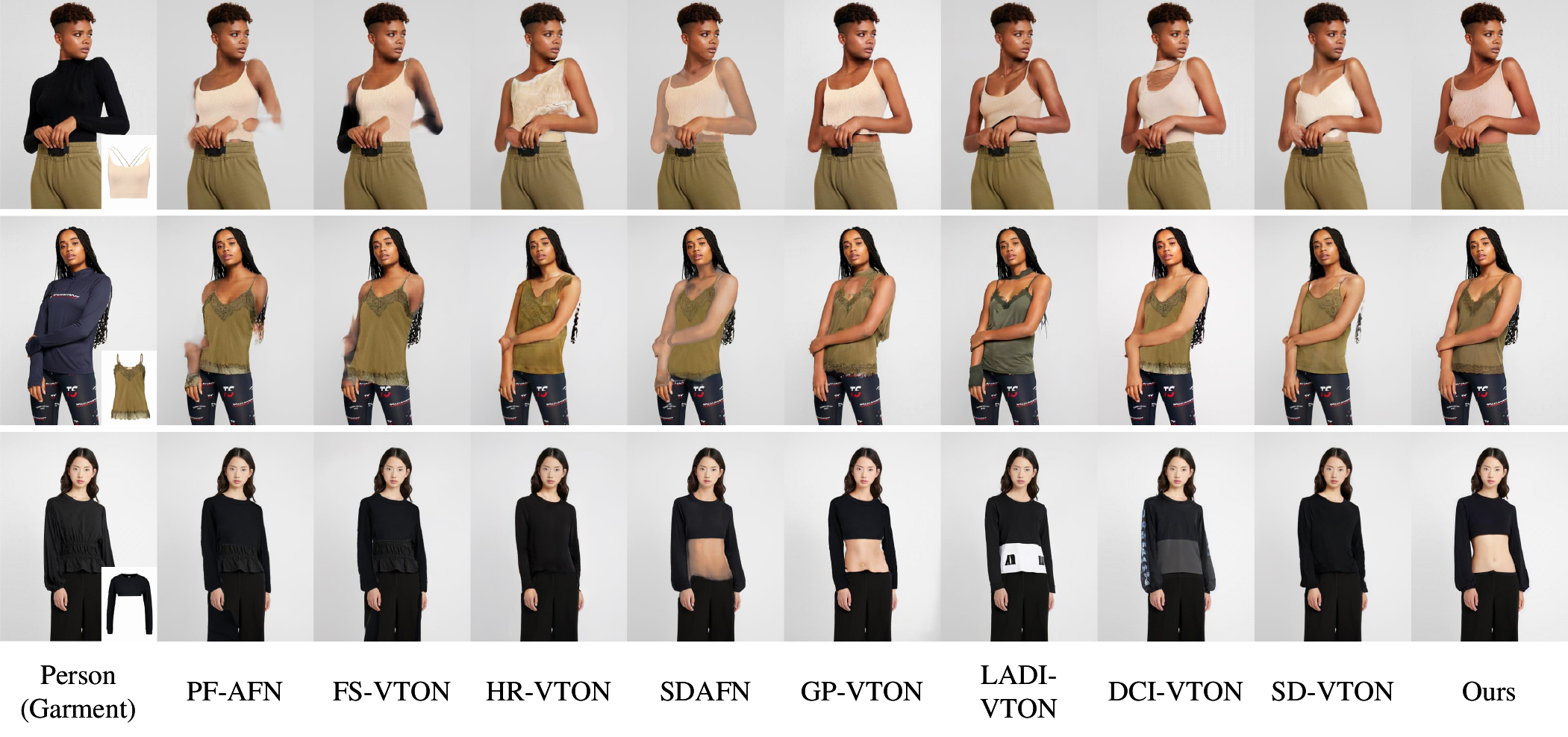}	
	\caption{Qualitative comparison on VITON-HD dataset~\cite{choi2021viton}. Please zoom in for a better view.} 
	\label{fig:vitonhd}
\end{figure}

\begin{figure}
	\centering
	\includegraphics[width=\textwidth]{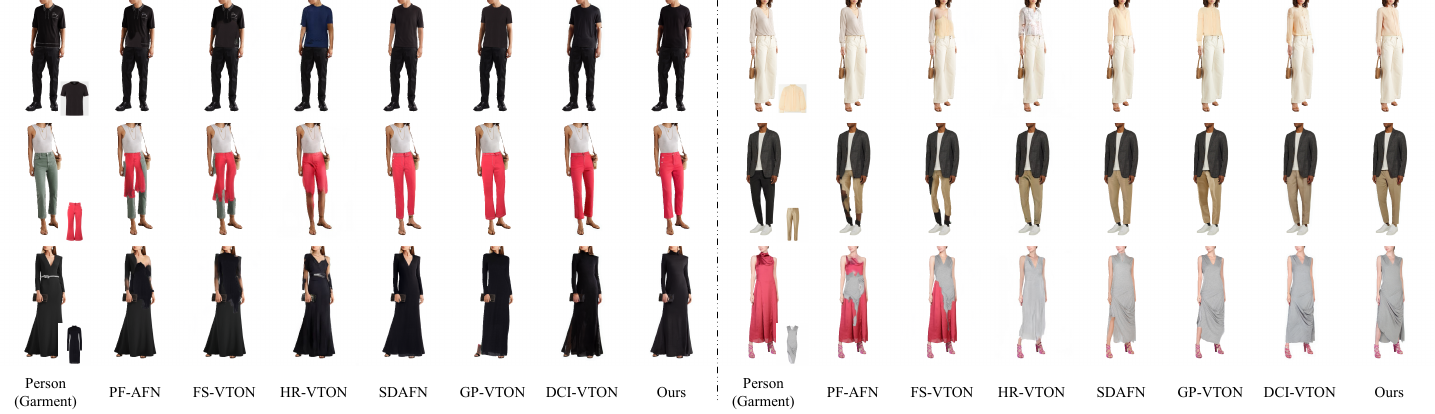}	
	\caption{Qualitative comparison on DressCode~\cite{morelli2022dress}. Categories in each row from top to bottom are upper, lower, and dresses, respectively. Please zoom in for a better view.} 	
	\label{fig:dresscode}
\end{figure}

\begin{figure}[t]
	\begin{floatrow}
		\capbtabbox{
			\resizebox{0.48\textwidth}{!}{
				\begin{tabular}{llllll}
					\toprule
					Variant & FID$_u$$\downarrow$ & FID$_p$$\downarrow$ & PSNR$\uparrow$ & SSIM$\uparrow$ & LPIPS$\downarrow$ \\
					\midrule
					$\dagger$Vanilla & - & 16.093 & 23.68 & 0.884 & 0.097 \\
					$\dagger$w/ fixed grouping & - & \underline{14.941} & \underline{24.96} & \underline{0.895} & \underline{0.087} \\
					$\dagger$w/ DSDM & - & \textbf{13.902} & \textbf{25.18} & \textbf{0.901} & \textbf{0.083} \\
					\midrule
					$\star$Vanilla & 8.754 & 5.521 & 24.01 & 0.882 & 0.080 \\
					$\star$w/ DSDM & \underline{8.639} & \underline{5.079} & 24.45 & 0.888 & 0.071 \\
					$\star$w/ DSDM \& DITP- & 9.032 & 5.271 & \textbf{24.97} & \textbf{0.893} & \textbf{0.062}\\
					$\star$w/ DSDM \& DITP & \textbf{8.530} & \textbf{4.845} & \underline{24.71} & \underline{0.892} & \underline{0.065} \\
					\bottomrule
			\end{tabular}}
		}{
			\vspace{2mm}
			\caption{Ablation study on VITON-HD dataset~\cite{choi2021viton}. Variants marked with $\dagger$ are evaluated on the warped results, and those marked with $\star$ are evaluated on the try-on results. The best results are in \textbf{bold}, and the second best are \underline{underlined}.}
			\label{tab:ablation}
		}
		\ffigbox{
			\includegraphics[width=0.45\textwidth]{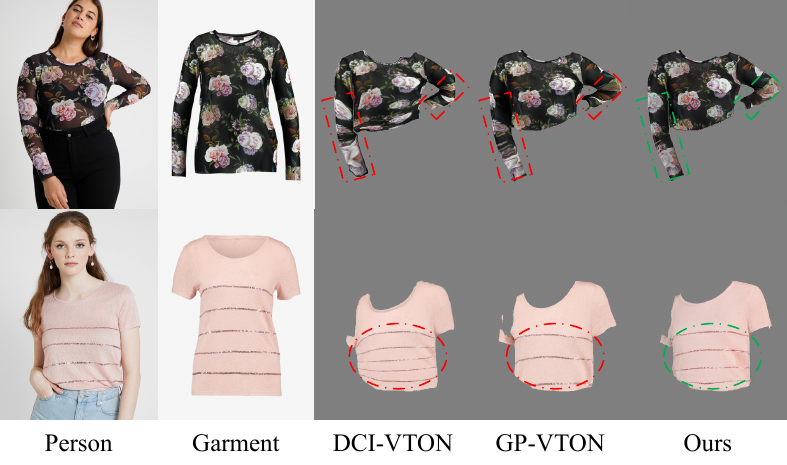}
		}{
			\vspace{-2mm}
			\caption{Visual comparison of warped garments. The texture within the red box appears stretched or squeezed.}
			\label{fig:warp_compare}
		}
	\end{floatrow}
\end{figure}

Tab.~\ref{tab:dresscode} summarizes the quantitative comparison of D$^4$-VTON with other methods on the DressCode dataset. For DressCode-Upper and DressCode-Lower, our method achieves superior results across all metrics. For DressCode-Dresses, D$^4$-VTON outperforms other methods on all unpaired metrics. In the paired setting, D$^4$-VTON shows the best performance on most metrics, except for FID$_p$ and KID$_p$, where it is slightly outperformed by SDAFN and GP-VTON. This may suggest that SDAFN and GP-VTON are overfitting, resulting in better paired performance but weaker unpaired results.\\
\textbf{Qualitative Results.} We also qualitatively compared our approach with other methods on the VITON-HD and DressCode datasets, as shown in Fig.~\ref{fig:vitonhd} and Fig.~\ref{fig:dresscode}. It can be observed that GAN-based methods struggle to generate realistic human body parts, such as arms or abdomen. SDAFN, a purely CNN-based method, performs well on the DressCode dataset with simple poses but struggles with scenarios lacking detailed body shape and skin color information, such as the VITON-HD dataset, leading to unconvincing results. GP-VTON effectively preserves garment textures by learning masks, but this approach results in noticeable "copy and paste" artifacts. Among diffusion methods, LADI-VTON fails to capture the accurate texture of the target garment due to misleading textual information, producing try-on results that differ significantly from the target garment. Although DCI-VTON incorporates the garment image as a condition, it fails to distinguish the garment's semantics, resulting in locally distorted outputs. In comparison, D$^4$-VTON applies multiple local flows via the dynamic semantics disentangling technique, resulting in more accurate warped results. Besides, the design of DITP helps D$^4$-VTON narrow the gap between incomplete and complete latents, alleviating the difficulty of restoring multiple degradations. This leads to try-on results that are more realistic and consistent with garment types.\\
\textbf{Ablation Study.} To demonstrate the effectiveness of our DSDM, we conducted experiments using three different settings to train the deformation network: Vanilla, with fixed grouping, and with DSDM. In the fixed grouping setting, we evenly divided all channels sequentially without using the Dynamic Semantics Selector. The results, shown in the upper part of Tab.~\ref{tab:ablation}, demonstrate significant improvements, indicated by the symbol $\dagger$ on metrics evaluated on warped garment items, even with the simple fixed grouping approach. This validates the effectiveness of our semantics grouping design. With the full incorporation of DSDM, we achieve the most accurate and perceptually friendly deformation results.

\begin{figure}
	\centering
	\includegraphics[width=\linewidth]{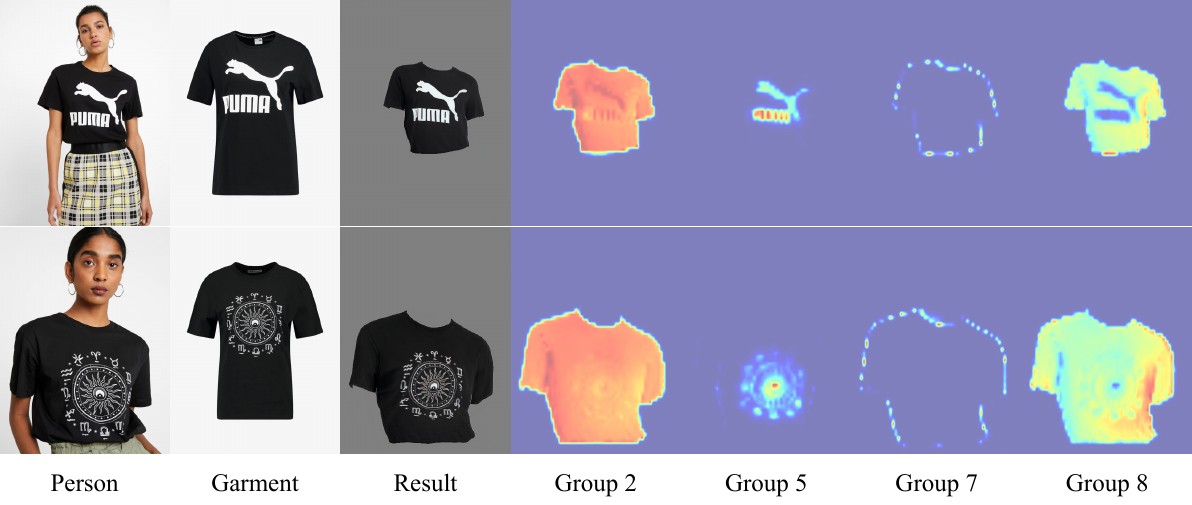}
	\vspace{-8mm}
	\caption{Visualization of different groups of the attention map generated by the final DSDM. Each map demonstrates distinct semantic responses.}
	\label{fig:group_attn}
\end{figure}

We also present the evaluation of the final try-on results in the lower part of Tab.~\ref{tab:ablation}, with variants marked by the symbol $\star$. Upon introducing DSDMs into the vanilla baseline, we observe significant improvements across all metrics, thereby demonstrating the effectiveness of our DSDM again. Importantly, even without explicit supervision for optimizing flows specific to semantics, our DSDM effectively implements meaningful semantics grouping. This is illustrated in Fig.~\ref{fig:group_attn}, which visualizes the fine attention map $a_f$ from the last DSDM for several groups. It is evident that channels with similar semantic information are grouped together by the DSDM, while those with different semantic information are separated for local warping. Specifically, Group 2 captures global garment information, Group 5 identifies patterns, Group 7 focuses on garment contour information, and Group 8 includes supplementary details. We further provide a qualitative comparison of warped results against two representative methods, DCI-VTON and GP-VTON, in Fig.~\ref{fig:warp_compare}. DCI-VTON exhibits texture squeezing and stretching phenomena, whereas GP-VTON mitigates texture squeezing using dynamic gradient truncation but still experiences stretching in the sleeve area. In contrast, our method effectively avoids these issues.

On the other hand, the effectiveness of the proposed DITP can also be observed in the lower part of Tab.~\ref{tab:ablation}. The setting ``w/ DSDM \& DITP-'' indicates using DSDM while removing the denoising path, whereas ``w/ DSDM \& DITP'' represents our full model. We observed that the former setting improves performance in the paired setting but notably underperforms in the unpaired setting. This emphasizes the importance of the denoising path in enhancing the generality of the entire framework, as demonstrated in the latter setting. More importantly, both settings incorporating DITP show substantial improvements compared to other variants, validating its effectiveness. Additional qualitative ablation results are provided in the supplementary materials.\vspace{-2mm}
\vspace{-8mm}
\vspace{-8mm}
\section{Limitation}
\begin{figure}
	\centering
	\includegraphics[width=\linewidth]{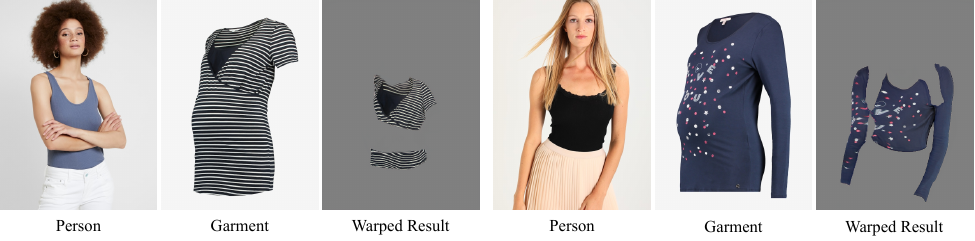}
	\caption{Visualization of failure cases.}
	\label{fig:limitation}
\end{figure}

Despite D$^4$-VTON demonstrating powerful capabilities in virtual try-on tasks, there are still limitations, as illustrated by some failure cases in Fig.~\ref{fig:limitation}. When dealing with garments that are partially obscured from view, such as the right sleeve, our DSDM may struggle to understand the missing semantic components. Consequently, it may generate incorrect shapes in the warped garment or incorrectly stretch or replicate textures from other semantic areas, such as the torso. Addressing these challenges remains a focus for our future work.
\vspace{-8mm}

\section{Conclusion}
In this paper, we propose a novel virtual try-on method, D$^4$-VTON, which primarily incorporates two key techniques: the Dynamic Semantics Disentangling Module and Differential Information Tracking. While the former leverages self-discovered semantics aggregation to enhance garment warping, the latter captures differential information to mitigate learning ambiguities caused by restoration for multiple degradations in a diffusion-based virtual try-on framework with negligible overhead. Comprehensive experiments validate the effectiveness of these techniques in both quantitative and qualitative evaluations.

\section*{Acknowledgement}
This project is supported by the National Natural Science Foundation of China (62102381, 41927805); Shandong Natural Science Foundation (ZR2021QF035); the National Key R\&D Program of China (2022ZD0117201); and the China Postdoctoral Science Foundation (2020M682240, 2021T140631).

\par\vfill\par
\bibliographystyle{splncs04}
\bibliography{main}

\begin{thebibliography}{10}
\providecommand{\url}[1]{\texttt{#1}}
\providecommand{\urlprefix}{URL }
\providecommand{\doi}[1]{https://doi.org/#1}

\bibitem{bai2022single}
Bai, S., Zhou, H., Li, Z., Zhou, C., Yang, H.: Single stage virtual try-on via
  deformable attention flows. In: ECCV. pp. 409--425. Springer (2022)

\bibitem{binkowski2018demystifying}
Bi{\'n}kowski, M., Sutherland, D.J., Arbel, M., Gretton, A.: Demystifying mmd
  gans. In: ICLR (2018)

\bibitem{bookstein1989principal}
Bookstein, F.L.: Principal warps: Thin-plate splines and the decomposition of
  deformations. IEEE TPAMI  \textbf{11}(6),  567--585 (1989)

\bibitem{chen2023size}
Chen, C.Y., Chen, Y.C., Shuai, H.H., Cheng, W.H.: Size does matter: Size-aware
  virtual try-on via clothing-oriented transformation try-on network. In: ICCV.
  pp. 7513--7522 (2023)

\bibitem{choi2021viton}
Choi, S., Park, S., Lee, M., Choo, J.: Viton-hd: High-resolution virtual try-on
  via misalignment-aware normalization. In: CVPR. pp. 14131--14140 (2021)

\bibitem{Du2023}
Du, Y., Zhan, J., He, S., Li, X., Dong, J., Chen, S., Yang, M.H.: One-for-all:
  Towards universal domain translation with a single stylegan. arXiv preprint
  arXiv:2310.14222  (2023)

\bibitem{fele2022c}
Fele, B., Lampe, A., Peer, P., Struc, V.: C-vton: Context-driven image-based
  virtual try-on network. In: WACV. pp. 3144--3153 (2022)

\bibitem{ge2021parser}
Ge, Y., Song, Y., Zhang, R., Ge, C., Liu, W., Luo, P.: Parser-free virtual
  try-on via distilling appearance flows. In: CVPR. pp. 8485--8493 (2021)

\bibitem{goodfellow2014generative}
Goodfellow, I., Pouget-Abadie, J., Mirza, M., Xu, B., Warde-Farley, D., Ozair,
  S., Courville, A., Bengio, Y.: Generative adversarial nets. In: NeurIPS.
  vol.~27 (2014)

\bibitem{gou2023taming}
Gou, J., Sun, S., Zhang, J., Si, J., Qian, C., Zhang, L.: Taming the power of
  diffusion models for high-quality virtual try-on with appearance flow. In:
  ACM MM. pp. 7599--7607 (2023)

\bibitem{han2019clothflow}
Han, X., Hu, X., Huang, W., Scott, M.R.: Clothflow: A flow-based model for
  clothed person generation. In: ICCV. pp. 10471--10480 (2019)

\bibitem{han2018viton}
Han, X., Wu, Z., Wu, Z., Yu, R., Davis, L.S.: Viton: An image-based virtual
  try-on network. In: CVPR. pp. 7543--7552 (2018)

\bibitem{he2022style}
He, S., Song, Y.Z., Xiang, T.: Style-based global appearance flow for virtual
  try-on. In: CVPR. pp. 3470--3479 (2022)

\bibitem{ho2020denoising}
Ho, J., Jain, A., Abbeel, P.: Denoising diffusion probabilistic models. In:
  NeurIPS. vol.~33, pp. 6840--6851 (2020)

\bibitem{jang2016categorical}
Jang, E., Gu, S., Poole, B.: Categorical reparameterization with
  gumbel-softmax. arXiv preprint arXiv:1611.01144  (2016)

\bibitem{karras2019style}
Karras, T., Laine, S., Aila, T.: A style-based generator architecture for
  generative adversarial networks. In: CVPR. pp. 4401--4410 (2019)

\bibitem{kingma2013auto}
Kingma, D.P., Welling, M.: Auto-encoding variational bayes. arXiv preprint
  arXiv:1312.6114  (2013)

\bibitem{lee2022high}
Lee, S., Gu, G., Park, S., Choi, S., Choo, J.: High-resolution virtual try-on
  with misalignment and occlusion-handled conditions. In: ECCV. pp. 204--219.
  Springer (2022)

\bibitem{li2023virtual}
Li, Z., Wei, P., Yin, X., Ma, Z., Kot, A.C.: Virtual try-on with pose-garment
  keypoints guided inpainting. In: ICCV. pp. 22788--22797 (2023)

\bibitem{li2023grouplane}
Li, Z., Han, C., Ge, Z., Yang, J., Yu, E., Wang, H., Zhao, H., Zhang, X.:
  Grouplane: End-to-end 3d lane detection with channel-wise grouping. In: ICLR
  (2024)

\bibitem{lin2017feature}
Lin, T.Y., Doll{\'a}r, P., Girshick, R., He, K., Hariharan, B., Belongie, S.:
  Feature pyramid networks for object detection. In: CVPR. pp. 2117--2125
  (2017)

\bibitem{maddison2016concrete}
Maddison, C.J., Mnih, A., Teh, Y.W.: The concrete distribution: A continuous
  relaxation of discrete random variables. arXiv preprint arXiv:1611.00712
  (2016)

\bibitem{morelli2023ladi}
Morelli, D., Baldrati, A., Cartella, G., Cornia, M., Bertini, M., Cucchiara,
  R.: Ladi-vton: Latent diffusion textual-inversion enhanced virtual try-on.
  In: ACM MM. pp. 8580--8589 (2023)

\bibitem{morelli2022dress}
Morelli, D., Fincato, M., Cornia, M., Landi, F., Cesari, F., Cucchiara, R.:
  Dress code: High-resolution multi-category virtual try-on. In: CVPR. pp.
  2231--2235 (2022)

\bibitem{parmar2022aliased}
Parmar, G., Zhang, R., Zhu, J.Y.: On aliased resizing and surprising subtleties
  in gan evaluation. In: CVPR. pp. 11410--11420 (2022)

\bibitem{rombach2022high}
Rombach, R., Blattmann, A., Lorenz, D., Esser, P., Ommer, B.: High-resolution
  image synthesis with latent diffusion models. In: CVPR. pp. 10684--10695
  (2022)

\bibitem{shim2023towards}
Shim, S.H., Chung, J., Heo, J.P.: Towards squeezing-averse virtual try-on via
  sequential deformation. In: AAAI. vol.~38, pp. 4856--4863 (2024)

\bibitem{simonyan2014very}
Simonyan, K., Zisserman, A.: Very deep convolutional networks for large-scale
  image recognition. arXiv preprint arXiv:1409.1556  (2014)

\bibitem{sohl2015deep}
Sohl-Dickstein, J., Weiss, E., Maheswaranathan, N., Ganguli, S.: Deep
  unsupervised learning using nonequilibrium thermodynamics. In: ICML. pp.
  2256--2265. PMLR (2015)

\bibitem{Song2022}
Song, H., Du, Y., Xiang, T., Dong, J., Qin, J., He, S.: Editing out-of-domain
  gan inversion via differential activations. In: ECCV. pp. 1--17. Springer
  (2022)

\bibitem{song2020denoising}
Song, J., Meng, C., Ermon, S.: Denoising diffusion implicit models. In: ICLR
  (2020)

\bibitem{tang2023contrastive}
Tang, J., Zheng, G., Shi, C., Yang, S.: Contrastive grouping with transformer
  for referring image segmentation. In: CVPR. pp. 23570--23580 (2023)

\bibitem{Wang2018}
Wang, B., Zheng, H., Liang, X., Chen, Y., Lin, L., Yang, M.: Toward
  characteristic-preserving image-based virtual try-on network. In: ECCV. pp.
  589--604 (2018)

\bibitem{wang2004image}
Wang, Z., Bovik, A.C., Sheikh, H.R., Simoncelli, E.P.: Image quality
  assessment: from error visibility to structural similarity. IEEE TIP
  \textbf{13}(4),  600--612 (2004)

\bibitem{wei2023inferring}
Wei, Y., Ji, Z., Wu, X., Bai, J., Zhang, L., Zuo, W.: Inferring and leveraging
  parts from object shape for improving semantic image synthesis. In: CVPR. pp.
  11248--11258 (2023)

\bibitem{xie2023gp}
Xie, Z., Huang, Z., Dong, X., Zhao, F., Dong, H., Zhang, X., Zhu, F., Liang,
  X.: Gp-vton: Towards general purpose virtual try-on via collaborative
  local-flow global-parsing learning. In: CVPR. pp. 23550--23559 (2023)

\bibitem{xie2021towards}
Xie, Z., Huang, Z., Zhao, F., Dong, H., Kampffmeyer, M., Liang, X.: Towards
  scalable unpaired virtual try-on via patch-routed spatially-adaptive gan. In:
  NeurIPS. vol.~34, pp. 2598--2610 (2021)

\bibitem{Xu2021a}
Xu, C., Fu, Y., Liu, C., Wang, C., Li, J., Huang, F., Zhang, L., Xue, X.:
  Learning dynamic alignment via meta-filter for few-shot learning. In: CVPR.
  pp. 5182--5191 (2021)

\bibitem{Xu2021}
Xu, Y., Du, Y., Xiao, W., Xu, X., He, S.: From continuity to editability:
  Inverting gans with consecutive images. In: ICCV. pp. 13910--13918 (2021)

\bibitem{yang2023paint}
Yang, B., Gu, S., Zhang, B., Zhang, T., Chen, X., Sun, X., Chen, D., Wen, F.:
  Paint by example: Exemplar-based image editing with diffusion models. In:
  CVPR. pp. 18381--18391 (2023)

\bibitem{Yang2020}
Yang, H., Zhang, R., Guo, X., Liu, W., Zuo, W., Luo, P.: Towards
  photo-realistic virtual try-on by adaptively generating-preserving image
  content. In: CVPR. pp. 7850--7859 (2020)

\bibitem{zhang2018unreasonable}
Zhang, R., Isola, P., Efros, A.A., Shechtman, E., Wang, O.: The unreasonable
  effectiveness of deep features as a perceptual metric. In: CVPR. pp. 586--595
  (2018)

\bibitem{zhou2016view}
Zhou, T., Tulsiani, S., Sun, W., Malik, J., Efros, A.A.: View synthesis by
  appearance flow. In: ECCV. pp. 286--301. Springer (2016)

\bibitem{Zhou2022}
Zhou, Y., Xu, Y., Du, Y., Wen, Q., He, S.: Pro-pulse: Learning progressive
  encoders of latent semantics in gans for photo upsampling. IEEE TIP
  \textbf{31},  1230--1242 (2022)

\end{thebibliography}
\end{document}